\documentclass[11pt,a4paper]{article}

\usepackage[a4paper,top=22mm,bottom=24mm,left=25mm,right=25mm]{geometry}
\usepackage{fontspec}
\defaultfontfeatures{Ligatures=TeX}
\usepackage{microtype}
\usepackage{graphicx}
\usepackage{booktabs}
\usepackage{tabularx}
\usepackage{longtable}
\usepackage{array}
\usepackage{amsmath,amssymb}
\usepackage{enumitem}
\usepackage{xcolor}
\usepackage{tikz}
\usetikzlibrary{arrows.meta,positioning,calc,patterns}
\usepackage{pgfplots}
\pgfplotsset{compat=1.18}
\usepackage{titlesec}
\usepackage{fancyhdr}
\usepackage{caption}
\usepackage{float}
\usepackage{xurl}
\usepackage[numbers,sort&compress]{natbib}
\usepackage{hyperref}
\usepackage[nameinlink,noabbrev]{cleveref}
\usepackage{ragged2e}

\definecolor{EnemrayTeal}{HTML}{0B4B58}
\definecolor{EnemrayDeep}{HTML}{14373F}
\definecolor{EnemrayGold}{HTML}{D7B77A}
\definecolor{EnemrayGray}{HTML}{666666}
\definecolor{EnemrayRule}{HTML}{929292}
\definecolor{EnemrayPale}{HTML}{F7F4ED}

\hypersetup{
  colorlinks=true,
  linkcolor=EnemrayTeal,
  citecolor=EnemrayTeal,
  urlcolor=EnemrayTeal,
  pdftitle={Enemray: Toward Capable Language Models for Hassaniya},
  pdfauthor={Cheikh Ahmed}
}

\titleformat{\section}{\Large\bfseries\color{black}}{\thesection.}{0.55em}{}
\titleformat{\subsection}{\large\bfseries\color{black}}{\thesubsection}{0.55em}{}
\titleformat{\subsubsection}{\normalsize\bfseries\color{black}}{\thesubsubsection}{0.5em}{}
\titlespacing*{\section}{0pt}{1.55em}{0.55em}
\titlespacing*{\subsection}{0pt}{1.15em}{0.35em}
\titlespacing*{\subsubsection}{0pt}{0.95em}{0.25em}

\setlist[itemize]{leftmargin=1.45em,itemsep=0.16em,topsep=0.3em}
\renewcommand{\arraystretch}{1.16}
\newcommand{\model}{Enemray}
\definecolor{FigureInk}{HTML}{163E47}
\definecolor{FigureLine}{HTML}{A7B5B8}
\definecolor{FigureTeal}{HTML}{0B5966}
\definecolor{FigureTealPale}{HTML}{F0F6F6}
\definecolor{FigureGold}{HTML}{9B7435}
\definecolor{FigureGoldPale}{HTML}{FAF6EE}
\tikzset{
  fig arrow/.style={-{Latex[length=2.0mm,width=1.35mm]},draw=FigureTeal,line width=0.85pt},
  fig derived/.style={-{Latex[length=2.0mm,width=1.35mm]},draw=FigureGold,line width=0.85pt,dash pattern=on 3pt off 2pt},
  fig label/.style={font=\sffamily\fontsize{9}{11}\selectfont,text=FigureInk,align=center},
  fig small/.style={font=\sffamily\fontsize{7.6}{9.4}\selectfont,text=FigureInk!75,align=center},
  fig header/.style={font=\sffamily\fontsize{8}{10}\selectfont\bfseries,text=FigureInk,anchor=west},
  fig process/.style={rounded corners=2mm,draw=FigureTeal!35,fill=white,line width=0.65pt,minimum height=9mm,inner sep=2mm,align=center,font=\sffamily\fontsize{9}{11}\selectfont\bfseries,text=FigureInk},
  pics/model/.style={code={
    \draw[draw=FigureTeal,fill=white,line width=0.8pt,rounded corners=0.6mm] (-.32,-.25) rectangle (.32,.25);
    \draw[draw=FigureTeal,line width=0.75pt] (-.40,-.10)--(-.40,.33)--(.20,.33);
    \draw[draw=FigureTeal,line width=0.75pt] (-.48,-.02)--(-.48,.41)--(.12,.41);
    \foreach \x in {-.17,0,.17} \foreach \y in {-.12,.03} \fill[FigureTeal] (\x,\y) circle (.021);
  }},
  pics/document/.style={code={
    \draw[draw=FigureTeal,fill=white,line width=0.8pt,rounded corners=0.4mm] (-.26,-.32)--(.26,-.32)--(.26,.16)--(.10,.32)--(-.26,.32)--cycle;
    \draw[draw=FigureTeal,line width=0.7pt] (.10,.32)--(.10,.16)--(.26,.16);
    \foreach \y in {-.15,-.02,.11} \draw[draw=FigureTeal,line width=0.65pt] (-.16,\y)--(.08,\y);
  }},
  pics/books/.style={code={
    \draw[draw=FigureTeal,fill=white,line width=0.8pt,rounded corners=0.5mm] (-.33,-.29) rectangle (-.13,.30);
    \draw[draw=FigureTeal,fill=white,line width=0.8pt,rounded corners=0.5mm] (-.10,-.29) rectangle (.10,.30);
    \draw[draw=FigureTeal,fill=white,line width=0.8pt,rounded corners=0.5mm,rotate=12] (.13,-.29) rectangle (.33,.30);
    \draw[draw=FigureTeal,line width=0.65pt] (-.29,.17)--(-.17,.17) (-.06,.17)--(.06,.17);
  }},
  pics/clean/.style={code={
    \draw[draw=FigureTeal,line width=0.8pt,line cap=round,line join=round] (-.29,.24)--(.29,.24)--(.08,-.02)--(.08,-.23)--(-.08,-.30)--(-.08,-.02)--cycle;
  }},
  pics/curate/.style={code={
    \draw[draw=FigureGold,line width=0.8pt,fill=white,rounded corners=.5mm] (-.25,-.30) rectangle (.23,.30);
    \draw[draw=FigureGold,line width=0.65pt] (-.14,.16)--(.11,.16) (-.14,.05)--(.04,.05);
    \draw[draw=FigureGold,line width=1pt,line cap=round,line join=round] (-.09,-.13)--(-.01,-.21)--(.29,.07);
  }},
  pics/replay/.style={code={
    \draw[draw=FigureGold,line width=.8pt,-{Latex[length=1.4mm]}] (25:.22) arc (25:330:.22);
  }},
  pics/corpus/.style={code={
    \draw[draw=FigureTeal,fill=white,line width=0.8pt] (-.36,.24)--(-.36,-.23) arc (180:360:.36 and .12)--(.36,.24);
    \draw[draw=FigureTeal,line width=0.65pt] (-.36,-.01) arc (180:360:.36 and .12);
    \draw[draw=FigureTeal,fill=white,line width=0.8pt] (0,.24) ellipse (.36 and .12);
  }}
}

\begin{document}

\thispagestyle{empty}

\vspace*{3mm}
\noindent
\raisebox{-0.25\height}{\includegraphics[height=12mm]{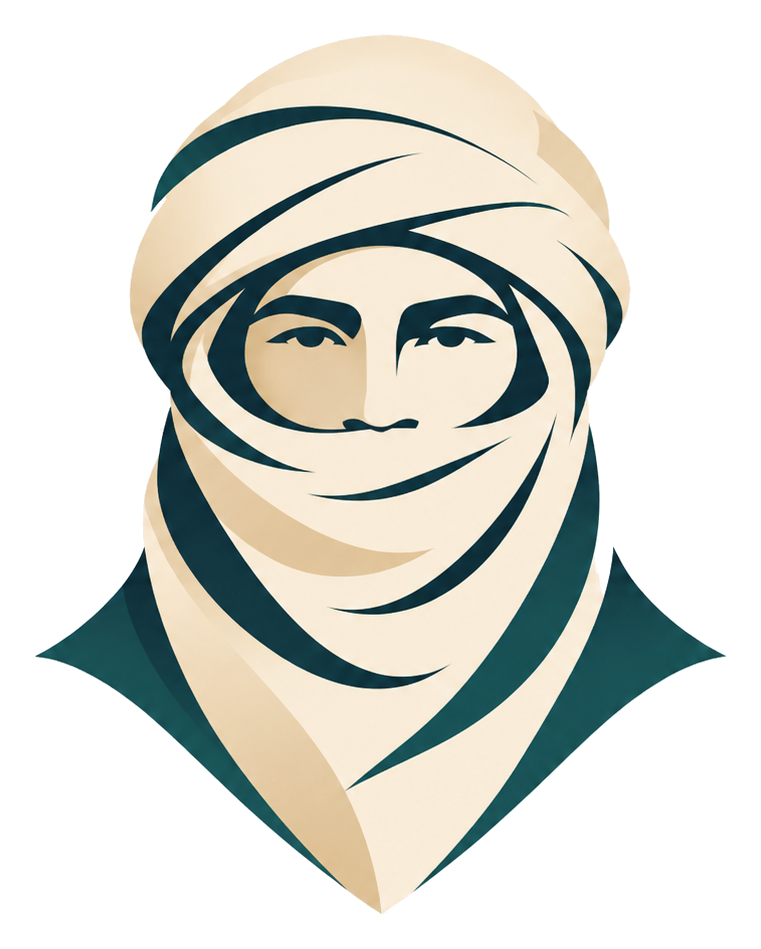}}\hspace{2.2mm}%
{\sffamily\fontsize{16.5}{18}\selectfont\bfseries\color{EnemrayTeal}Enemray}

\vspace{3mm}
\noindent\color{EnemrayRule}\rule{\textwidth}{0.55pt}\color{black}

\vspace{7mm}
\begin{center}
{\fontsize{19}{23}\selectfont\bfseries
Enemray:\\[1.2mm]
Toward Capable Language Models for Hassaniya\par}
\vspace{5mm}
{\normalsize Cheikh Ahmed\par}
\end{center}

\vspace{5mm}
\begin{center}
{\Large\bfseries Abstract}
\end{center}
\vspace{1mm}

\noindent
We introduce \model{}, a Hassaniya-centric language model that enables general-purpose interaction in Hassaniya. Enemray is trained around a stability--plasticity objective: acquire strong Hassaniya linguistic and cultural competence while preserving the general reasoning, multilingual, instruction-following, and safety behaviors of a capable instruction-tuned model. The development pipeline separates language acquisition from behavioral specialization. A separately assembled continual-pretraining corpus provides broad exposure to natural Hassaniya and Mauritanian text; layer-selective continual pretraining learns a compact language-specific parameter update; that update is transferred into the instruction-tuned parameter space; and supervised post-training develops conversational, cultural, literary, task-oriented, and cross-lingual behavior. The supervised corpus integrates selected public Hassaniya and Mauritanian resources with a substantially larger body of newly collected, reconstructed, curated, and constructed instruction data, while policy-generated replay provides a retention signal from the reference model's own behavior distribution. The resulting collection is substantially larger and broader in purpose than existing Hassaniya text resources. In evaluation, Enemray achieves the strongest English $\rightarrow$ Hassaniya translation among the compared open and proprietary models and the highest overall score on Mauritanian translation error detection, while retaining most of the general capabilities of its instruction-tuned base model on mathematical reasoning, knowledge, code generation, and function calling. This report describes the motivation, data construction, model design, training methodology, and evaluation of Enemray.

\vspace{3mm}
\noindent{\small\textbf{Keywords:} Hassaniya Arabic; low-resource language modeling; continual pretraining; language adaptation.}

\begin{figure}[H]
\centering
\begin{minipage}[c]{0.58\textwidth}
\centering
% Translation results bar chart (DeepSeek-style: our model in hatched blue, baselines muted).
% Sized for the left panel (0.58\textwidth) of the first-page figure; all scores are labeled.
% Scores: eval/results. Use "nan" for a model whose results are not available yet.
\definecolor{BarOurs}{HTML}{4868F8}
\definecolor{BarBase}{HTML}{A8C0F8}
\definecolor{BarGPT}{HTML}{E8D0A0}
\definecolor{BarGemini}{HTML}{C9C9C9}
\definecolor{BarQwen}{HTML}{EFE3C6}
\tikzset{
  trvalue/.style={font=\sffamily\fontsize{4.3}{4.5}\selectfont, text=black!70, rotate=90, anchor=west, inner sep=0.8pt,
    /pgf/number format/fixed, /pgf/number format/fixed zerofill, /pgf/number format/precision=1, /pgf/number format/assume math mode=true},
}
\begin{tikzpicture}
\begin{axis}[
  width=1.04\textwidth,
  height=4.9cm,
  ybar=0.6pt,
  bar width=7pt,
  ymin=0, ymax=60,
  ytick={0,20,40,60},
  symbolic x coords={EH-spBLEU,EH-chrF,HE-spBLEU,HE-chrF},
  xtick=data,
  xticklabels={{spBLEU},{chrF++},{spBLEU},{chrF++}},
  xticklabel style={font=\sffamily\footnotesize},
  enlarge x limits=0.12,
  ymajorgrids,
  grid style={draw=black!10},
  axis line style={draw=black!40},
  tick style={draw=black!40},
  xtick pos=lower,
  yticklabel style={font=\sffamily\scriptsize, /pgf/number format/assume math mode=true},
  unbounded coords=discard,
  legend style={draw=none, font=\sffamily\fontsize{6.5}{7.5}\selectfont, legend columns=3, column sep=6pt,
    at={(0.5,1.03)}, anchor=south},
  legend image code/.code={\draw[#1, draw=none] (0cm,-0.08cm) rectangle (0.28cm,0.12cm);},
  clip=false,
]
\addplot[fill=BarBase, draw=none, nodes near coords, nodes near coords style={trvalue}]
  coordinates {(EH-spBLEU,8.83) (EH-chrF,26.09) (HE-spBLEU,21.94) (HE-chrF,42.91)};
\addlegendentry{Gemma 4 E4B-it}
\addplot[fill=BarGPT, draw=none, nodes near coords, nodes near coords style={trvalue}]
  coordinates {(EH-spBLEU,8.61) (EH-chrF,26.14) (HE-spBLEU,26.95) (HE-chrF,48.76)};
\addlegendentry{GPT-5.6 Luna}
\addplot[fill=BarGemini, draw=none, nodes near coords, nodes near coords style={trvalue}]
  coordinates {(EH-spBLEU,5.32) (EH-chrF,21.59) (HE-spBLEU,28.31) (HE-chrF,49.93)};
\addlegendentry{Gemini 3.5 Flash Lite}
\addplot[fill=BarQwen, draw=none, nodes near coords, nodes near coords style={trvalue}]
  coordinates {(EH-spBLEU,8.30) (EH-chrF,25.34) (HE-spBLEU,20.62) (HE-chrF,44.18)};
\addlegendentry{Qwen 3.8 Flash}
\addplot[fill=BarOurs, draw=none, postaction={pattern=north east lines, pattern color=white},
  nodes near coords, nodes near coords style={trvalue, font=\sffamily\fontsize{4.3}{4.5}\selectfont\bfseries, text=black}]
  coordinates {(EH-spBLEU,12.64) (EH-chrF,29.42) (HE-spBLEU,29.52) (HE-chrF,48.99)};
\addlegendentry{Enemray}
\node[font=\sffamily\scriptsize, text=black!70, anchor=north] at (rel axis cs:0.285,-0.15) {English $\rightarrow$ Hassaniya};
\node[font=\sffamily\scriptsize, text=black!70, anchor=north] at (rel axis cs:0.715,-0.15) {Hassaniya $\rightarrow$ English};
\end{axis}
\end{tikzpicture}
\end{minipage}\hfill
\begin{minipage}[c]{0.39\textwidth}
\centering
\includegraphics[width=\linewidth]{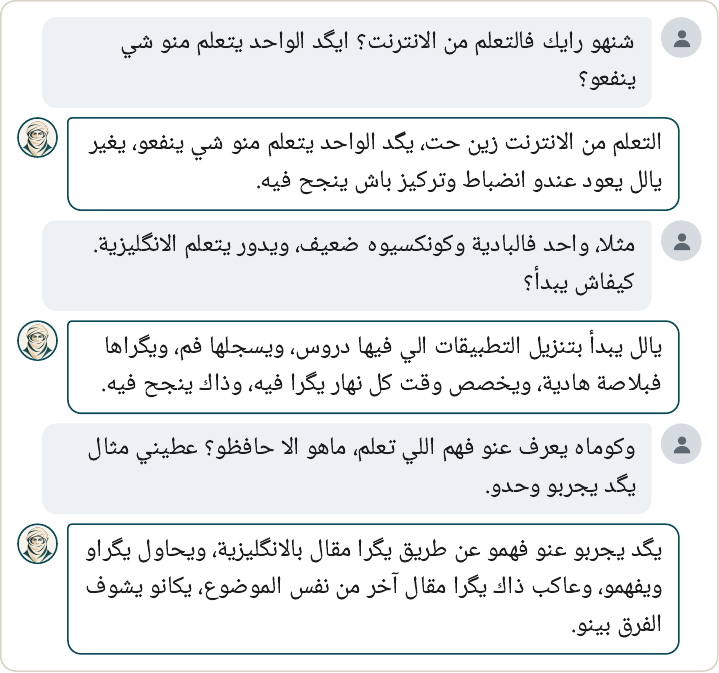}
\end{minipage}
\caption{Left: English $\leftrightarrow$ Hassaniya translation on the Alexandria (Mauritania) test set, compared with open and proprietary models. Right: a three-turn conversation with Enemray in Hassaniya (unedited model output).}
\label{fig:translation}
\end{figure}

\vspace{7mm}
\section{Introduction}
\label{sec:intro}

\subsection{Why Hassaniya}

Hassaniya (\textit{\d{H}ass\={a}niyya}) is a western Arabic variety spoken across a contiguous region of the western Sahara and Sahel, with its largest speaker population in Mauritania, where it functions broadly as a lingua franca. Its historical development reflects sustained contact with Classical and Modern Standard Arabic, Berber languages, and, in modern usage, French and other regional languages \cite{tainecheikh2020,glottologhassaniyya}. These contact dynamics are visible in Hassaniya phonology, morphology, lexicon, register, and patterns of code-switching, making it a linguistically distinct modeling target \cite{tainecheikh2020}.

Hassaniya also carries a substantial oral and literary tradition. UNESCO's documentation of the Moorish epic T'heydinn describes a body of dozens of poems in Hassaniya that preserves collective memory, social values, and historical narratives \cite{unesco_t_heydinn}. This cultural role motivates a model spanning everyday conversation and translation alongside local forms of expression, heritage knowledge, and poetic structure.

Despite this linguistic and cultural breadth, Hassaniya remains sparsely represented in computational resources. Earlier large-scale Arabic dialect identification work excluded Mauritania because sufficient dialectal Twitter data could not be collected \cite{abdelali2020qadi}. More recent resources have expanded coverage through sentiment data, English--Hassaniya parallel corpora, community datasets, culturally grounded Arabic instruction, multi-country dialectal translation, task-oriented dialogue, proverb collections, OCR text, and speech \cite{elarby2025hassaniya,boughale2026smart9k,hassania_dah,alwajih2025palm,elmekki2026alexandria,aiforrim,amthalhassaniya,hassaniya_stories,talafha2024casablanca}. The resulting ecosystem is valuable, but it is distributed across distinct tasks and modalities rather than organized as a broad training distribution for a general-purpose Hassaniya language model.

The central objective of \model{} is to make Hassaniya a first-class interaction language for a general-purpose language model: a language in which the model can reason, explain, translate, converse, write, and engage with culturally specific knowledge and expressive forms.

\subsection{The adaptation problem}

Specializing a strong multilingual model to an underrepresented language creates a stability--plasticity problem. The model must acquire new linguistic structure from a comparatively concentrated target-language corpus, while retaining capabilities learned from much broader pretraining and post-training distributions. Naive continued training can improve target-language modeling while moving parameters away from behaviors that support reasoning, instruction following, multilinguality, coding, and safety. This challenge is particularly acute for assistants that must combine target-language competence with broad general-purpose reasoning and interaction capabilities.

Enemray addresses this problem by separating \emph{language acquisition} from \emph{behavioral specialization}. First, a language-specific update is learned from a base model through layer-selective continual pretraining (CPT). Second, this update is transferred into the corresponding instruction-tuned model through parameter-space composition. Third, supervised post-training, implemented as supervised fine-tuning (SFT), binds the acquired Hassaniya competence to interaction behaviors. General-capability replay is included within the third stage to maintain exposure to broader assistant behavior during supervised post-training. Enemray implements this decomposition using Gemma 4 E4B. The design follows the broader observation that language addition, parameter-efficient adaptation, and post-training retention are related but distinct optimization problems \cite{owodunni2025layra,biderman2024lora,chen2026opr}.

\subsection{Contributions}

The technical contributions of this report are fourfold:
\begin{itemize}
    \item \textbf{Large-scale Hassaniya data construction.} We construct two complementary data streams: a separately assembled continual-pretraining corpus for broad language acquisition, and a Hassaniya supervised corpus that integrates selected public resources with a substantially larger body of Enemray-constructed instruction data. Together they cover naturally occurring language, Mauritanian journalism and cultural writing, poetry, translation, grammar, role-play, task-oriented dialogue, and cultural knowledge. The resulting collection supports a broader training objective than the individual task-specific resources summarized in this report.
    \item \textbf{Layer-selective language acquisition.} We use LayRA-style continual pretraining to place the primary language-learning update in selected early and late layers, reducing the number of trainable parameters while keeping intermediate-layer weights fixed \cite{owodunni2025layra}.
    \item \textbf{Language capability transfer.} We represent the acquired Hassaniya competence as a parameter update and compose it with the corresponding instruction-tuned model. This follows the task-vector view that fine-tuning-induced parameter differences can encode transferable behavior in weight space \cite{ilharco2022taskarithmetic}.
    \item \textbf{Replay-regularized post-training.} We combine Hassaniya-centric supervised learning with offline general-capability replay. The replay component is motivated by recent evidence that training on model-generated, policy-aligned trajectories can reduce forgetting during continual supervised fine-tuning \cite{chen2026opr}.
\end{itemize}

\begin{figure}[t]
\centering
\resizebox{\textwidth}{!}{\input{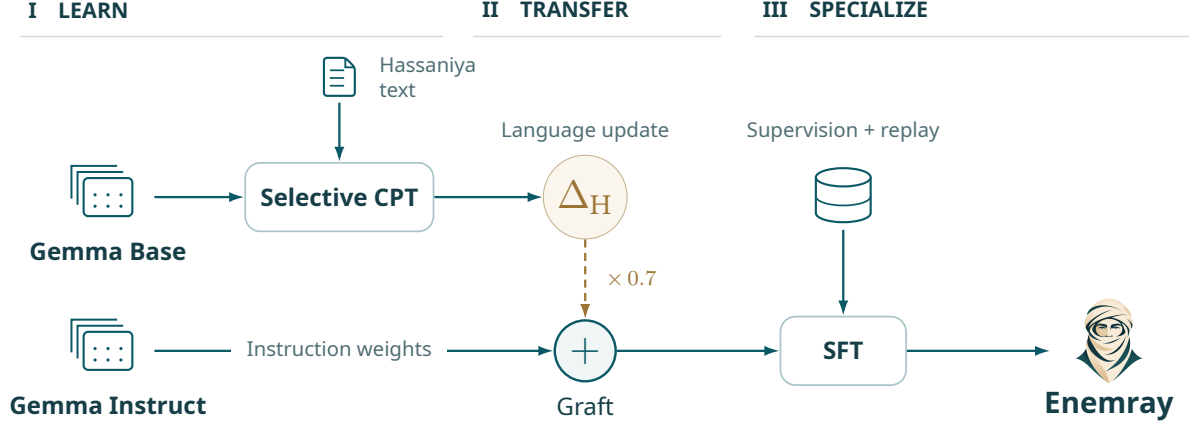}}
\caption{Enemray training pipeline. Selective continual pretraining learns a language update from natural Hassaniya text. The scaled update is added to the instruction-tuned checkpoint, which is then specialized using supervised examples and replay. Solid arrows indicate the main data or model flow; the dashed arrow denotes parameter transfer.}
\label{fig:pipeline}
\end{figure}

The full development path is summarized in Figure~\ref{fig:pipeline}. Each stage answers a different requirement: acquiring the language, transferring that acquisition to an interactive model, shaping Hassaniya behavior, and stabilizing broad capabilities during specialization.

\section{Model Design}
\label{sec:model}

\subsection{Architectural foundation}

Enemray adapts the Gemma 4 E4B base and instruction-tuned checkpoints. The backbone is a dense multimodal model with approximately 8 billion parameters including embeddings and approximately 4.5 billion effective parameters. The distinction reflects per-layer embedding tables rather than mixture-of-experts activation \cite{gemma4base}. The text decoder contains 42 layers, with hidden dimension 2,560, feed-forward dimension 10,240, and a vocabulary of 262,144 entries. The decoder repeats a block of five sliding-attention layers followed by one full-attention layer seven times, and the configured context capacity is 131,072 tokens \cite{gemma4config}.

\begin{table}[t]
\centering
\caption{Inherited Gemma 4 E4B configuration \cite{gemma4base,gemma4config}.}
\label{tab:model}
\small
\begin{tabularx}{0.94\textwidth}{@{}>{\bfseries}l X@{}}
\toprule
Property & Specification \\
\midrule
Parameter scale & Approximately 8B total; 4.5B effective \\
Text-decoder layers & 42 \\
Hidden / feed-forward dimension & 2,560 / 10,240 \\
Vocabulary & 262,144 entries \\
Attention heads / KV heads & 8 / 2 \\
Attention pattern & Seven blocks of five sliding and one full-attention layer \\
Attention-layer totals & 35 sliding and 7 full-attention layers \\
Sliding window & 512 tokens \\
Architectural context & 131,072 tokens \\
Per-layer embedding contribution & 256 dimensions \\
KV-state sharing & Final 18 text layers \\
Inherited input modalities & Text, image, audio \\
Adaptation modality & Text only \\
\bottomrule
\end{tabularx}
\end{table}

The complete multimodal model is loaded, while adaptation is restricted to attention and feed-forward projections in the text decoder. Vision and audio encoders, multimodal projectors, per-layer embedding components, token embeddings, and the language-model head remain frozen.

\subsection{Layer-selective adaptation}

Following LayRA, which concentrates language adaptation in the early and late layers of a transformer while leaving its intermediate representations unchanged \cite{owodunni2025layra}, Enemray restricts continual pretraining to the first ten and the final two layers of the text decoder:
\begin{equation}
\mathcal{L}=\{0,1,\ldots,9\}\cup\{40,41\}.
\end{equation}
Continual pretraining therefore updates 12 of the 42 decoder layers, and the remaining 30 intermediate layers retain their pretrained weights. Within each selected layer, low-rank adapters are attached to the query, key, value, and output projections of the attention block and to the up- and down-projections of the feed-forward block, while the feed-forward gate projection is left frozen. All adapters use rank $r=16$ and scaling parameter $\alpha=32$ \cite{hu2021lora}.

Continual pretraining thus adapts 72 weight matrices in total. During supervised post-training, adaptation is extended to every layer of the decoder and to all attention and feed-forward projections, including the gate projection, yielding 294 adapted weight matrices.

\subsection{Language capability transfer}

Let $\theta_{\mathrm{base}}$ denote the base checkpoint, $\theta_{\mathrm{post}}$ its instruction-tuned counterpart, and $\theta_{\mathrm{base,H}}$ the base checkpoint after Hassaniya continual pretraining. For each adapted weight matrix $W$, continual pretraining learns a low-rank update
\begin{equation}
\Delta W_{\mathrm{H}}=\frac{\alpha}{r}BA,
\end{equation}
where $A$ and $B$ are the learned low-rank factors. Because all other parameters remain unchanged, the language update of the full model is
\begin{equation}
\Delta_{\mathrm{H}}=\theta_{\mathrm{base,H}}-\theta_{\mathrm{base}},
\end{equation}
which equals $\Delta W_{\mathrm{H}}$ for each adapted matrix and zero elsewhere. This update is then added to the instruction-tuned checkpoint:
\begin{equation}
\theta_{\mathrm{H,post}}=\theta_{\mathrm{post}}+\gamma\Delta_{\mathrm{H}},
\qquad \gamma=0.7.
\label{eq:transfer}
\end{equation}
We refer to the resulting model $\theta_{\mathrm{H,post}}$ as the grafted checkpoint.

This formulation follows task arithmetic, which interprets differences between fine-tuned and pretrained weights as reusable directions in parameter space \cite{ilharco2022taskarithmetic}, and is closely related to chat-vector transfer, in which weight differences between base and instruction-tuned models are combined across checkpoints that share a common origin \cite{huang2024chatvector}. The transfer assumes that $\theta_{\mathrm{post}}$ is derived from $\theta_{\mathrm{base}}$ and shares its parameterization, so that an update learned relative to the base model remains meaningful in the neighborhood of the instruction-tuned weights. The coefficient $\gamma$ controls the strength of the transferred update; we use $\gamma=0.7$ as a fixed setting and leave its systematic selection to future work. Subsequent supervised training specializes the grafted checkpoint for Hassaniya interaction.

\section{Training Data}
\label{sec:data}

Data is the central constraint in Hassaniya language modeling. Public resources have expanded in recent years, but they remain distributed across narrow tasks such as sentiment classification, translation, cultural instruction, dialogue, speech, proverbs, and language learning. Enemray therefore treats the public ecosystem as a starting point rather than as a complete language-model training distribution. The resulting data pipeline separates two objectives: broad language acquisition from independently assembled natural text during continual pretraining, and behavioral specialization from a larger Hassaniya supervised mixture that incorporates selected public resources together with additional Enemray-constructed and curated data.

\subsection{Existing Hassaniya data landscape}
\label{sec:ecosystem}

Table~\ref{tab:ecosystem} summarizes publicly documented resources containing Hassaniya or explicitly labeled Mauritanian Arabic that we identified during corpus development. The resources are not directly comparable---they differ in modality, task, annotation, and unit of measurement---but together they show both the growing value and the fragmentation of the available ecosystem. The final column indicates which public resources are incorporated into Enemray; all of them are used in supervised post-training.

\begin{table}[tbp]
\centering
\caption{Publicly documented Hassaniya or Mauritanian-Arabic resources identified during corpus development. Corpus-wide sizes are shown when a source does not publish a Hassaniya-only count. ``SFT'' indicates incorporation into Enemray supervised post-training only.}
\label{tab:ecosystem}
\footnotesize
\renewcommand{\arraystretch}{1.1}
\begin{tabular}{@{}>{\bfseries\raggedright\arraybackslash}p{0.24\textwidth} >{\raggedright\arraybackslash}p{0.19\textwidth} >{\raggedright\arraybackslash}p{0.38\textwidth} >{\raggedright\arraybackslash}p{0.11\textwidth}@{}}
\toprule
Resource & Type & Publicly reported scale & Enemray use \\
\midrule
HASSANIYA \cite{elarby2025hassaniya} & Sentiment & 2,000 labeled Facebook comments & -- \\
HASSANIYA-DTCD \cite{elarby2025dtcd} & Classification & 1,851 labeled text records & -- \\
SMART-9K-H-CORP \cite{boughale2026smart9k} & Parallel text & 9,000 English--Hassaniya sentence pairs & -- \\
DAH \cite{hassania_dah} & Parallel text & 3,002 English--Hassaniya rows in Arabic script and Latin transliteration & SFT \\
PALM \cite{alwajih2025palm} & Cultural instruction & 1,298 Mauritania instructions, including 294 dialect-category instructions & SFT \\
Alexandria \cite{elmekki2026alexandria} & Multi-turn translation / dialogue & 107K samples overall across 13 Arab countries and 11 domains, including Mauritania & SFT \\
AI-for-RIM \cite{aiforrim} & Translation / dialogue & About 4,430 translation pairs and 594 support turns & SFT dialogue \\
Amthal Hassaniya \cite{amthalhassaniya} & Proverbs & 319 Hassaniya proverb examples with explanations & SFT \\
Hassaniya Stories OCR \cite{hassaniya_stories} & OCR / cultural text & 552 image--text rows & -- \\
Casablanca \cite{talafha2024casablanca} & Speech & About 48 hours overall across eight Arabic dialects, including Hassaniya & -- \\
Hassaniya Speech Dataset \cite{hassaniya_speech} & Speech & 294 audio--text rows & -- \\
Peace Corps Hassaniya material \cite{peacecorpshassaniya} & Language learning & Hassaniya--English instructional text and phrases & -- \\
Basic English--Hassaniya Dictionary \cite{hassaniyadictionary} & Lexical reference & More than 8,000 words, phrases, and sentences & -- \\
Hassaniya scripture resources \cite{scriptureearthmey,bibleparallel} & Parallel text & Verse-aligned Hassaniya scripture; edition-dependent & SFT \\
\bottomrule
\end{tabular}
\end{table}

The ecosystem is useful but task-specific. SMART-9K-H-CORP provides substantial translation supervision, Casablanca contributes speech coverage, and newer multidialectal efforts such as PALM and Alexandria include Mauritania within broader Arabic collections \cite{boughale2026smart9k,talafha2024casablanca,alwajih2025palm,elmekki2026alexandria}. None of these resources alone supplies the broad natural-text and general-purpose conversational distribution needed for a Hassaniya generative model. Enemray therefore integrates selected public supervision while constructing a substantially larger training distribution around natural Hassaniya text and general assistant behavior.

\subsection{Building the Enemray training corpus}
\label{sec:dataconstruction}

The Enemray training data comprise two complementary streams that serve distinct objectives. The first is a continual-pretraining corpus assembled from naturally occurring and long-form Hassaniya text, including social narratives, reconstructed long-form narratives, Mauritanian news and articles, cultural essays, poetry, and language-learning material. Its purpose is to expose the model to the distribution of the language itself. The second is a supervised corpus designed to shape assistant behavior. It integrates selected public resources from Table~\ref{tab:ecosystem} with a larger body of Enemray-constructed data spanning cultural knowledge, dialogue, translation, linguistic refinement, practical interaction, and Hassaniya literary forms.

Table~\ref{tab:dataoverview} summarizes the size of each stream after cleaning. Because the two streams are organized around different training units, the continual-pretraining corpus is measured in documents and the supervised corpus in conversations, and both are additionally reported in tokens of the Gemma 4 tokenizer. The supervised stage also includes a small general-capability replay set. Although it is not Hassaniya data, it is listed here to present the complete supervised mixture, and its role is discussed with the training methodology in Section~\ref{sec:training}.

\begin{table}[H]
\centering
\caption{Size of the Enemray training data after cleaning. Token counts are computed with the Gemma 4 tokenizer over message content, excluding system prompts and conversational formatting.}
\label{tab:dataoverview}
\small
\begin{tabularx}{\textwidth}{@{}X r r@{}}
\toprule
Component & Size & Gemma 4 tokens \\
\midrule
Continual-pretraining corpus & 11,629 documents & 20.9M \\
Hassaniya supervised data & 46,259 conversations & 3.0M \\
General-capability replay & 273 conversations & 0.4M \\
\midrule
Complete supervised mixture & 46,532 conversations & 3.4M \\
\bottomrule
\end{tabularx}
\end{table}

The supervised mixture contains 46,532 conversations with 50,273 assistant turns. Since each assistant turn is treated as a separate training example, the number of assistant turns corresponds to the number of supervised examples used during post-training, whereas the number of conversations reflects the size of the underlying dialogue collection.

Within this design, public datasets constitute one component of the supervised stage rather than its foundation: data constructed for Enemray account for roughly two thirds of the supervised response tokens, while public resources contribute the remaining third. In scale, the continual-pretraining corpus alone exceeds the individual public text resources in Table~\ref{tab:ecosystem}, which are typically measured in thousands of sentence pairs or instructions, by more than an order of magnitude.

\subsection{Data construction recipes}
\label{sec:datarecipes}

\begin{figure}[t]
\centering
\resizebox{\textwidth}{!}{\begin{tikzpicture}[x=1cm,y=1cm]
\path[use as bounding box] (0,-.12) rectangle (15.8,6.10);
\node[fig header] at (.12,5.79) {SOURCES};
\node[fig header] at (6.10,5.79) {PREPARATION};
\node[fig header] at (12.44,5.79) {TRAINING DATA};
\draw[FigureLine!55,line width=.5pt] (.15,5.43)--(5.08,5.43) (6.12,5.43)--(10.72,5.43) (12.46,5.43)--(15.60,5.43);

% Natural text provides language exposure and material for derived supervision.
\fill[FigureTealPale,rounded corners=2mm] (.14,3.75) rectangle (15.61,5.12);
\pic at (.85,4.43) {document};
\node[fig label,anchor=west,font=\sffamily\fontsize{9}{11}\selectfont\bfseries] at (1.43,4.64) {Natural Hassaniya};
\node[fig small,anchor=west] at (1.43,4.17) {Web\enspace /\enspace news\enspace /\enspace poetry};
\node[fig process,minimum width=3.26cm] (clean) at (8.27,4.43) {\hspace{6mm}Collect \& clean};
\pic[scale=.72] at (7.13,4.43) {clean};
\draw[fig arrow] (4.79,4.43)--(clean.west);
\pic at (13.85,4.50) {corpus};
\node[fig label,font=\sffamily\fontsize{9}{11}\selectfont\bfseries] at (13.85,3.94) {CPT corpus};
\draw[fig arrow] (clean.east)--(13.17,4.43);

% Public resources join the constructed supervision before the final SFT mixture.
\fill[FigureGoldPale,rounded corners=2mm] (.14,1.30) rectangle (15.61,2.66);
\pic at (.85,1.99) {books};
\node[fig label,anchor=west,font=\sffamily\fontsize{9}{11}\selectfont\bfseries] at (1.43,2.19) {Public resources};
\node[fig small,anchor=west] at (1.43,1.72) {Selected Hassaniya datasets};
\node[fig process,draw=FigureGold!45,minimum width=3.64cm] (curate) at (8.27,1.99) {\hspace{6mm}Construct \& curate};
\pic[scale=.72] at (6.74,1.99) {curate};
\draw[fig arrow] (4.79,1.99)--(curate.west);
\pic at (13.85,2.07) {corpus};
\node[fig label,font=\sffamily\fontsize{9}{11}\selectfont\bfseries] at (13.85,1.50) {SFT corpus};
\draw[fig arrow] (curate.east)--(13.17,1.99);

% The branch conveys shared source material without implying shared objectives.
\draw[fig derived,rounded corners=1.5mm] (2.88,3.75)--(2.88,3.18)--(8.27,3.18)--(curate.north);
\node[fig small,fill=white,inner sep=1.5mm,text=FigureGold] at (5.25,3.18) {Derived supervision};
\node[fig small] at (8.27,.95) {Translation\enspace /\enspace dialogue\enspace /\enspace correction};
\node[fig process,draw=FigureGold!45,fill=white,minimum height=6.5mm,minimum width=1.95cm,font=\sffamily\fontsize{8}{10}\selectfont] (replay) at (10.44,.30) {\hspace{4mm}Replay};
\pic[scale=.62] at (9.86,.30) {replay};
\draw[fig derived,rounded corners=1.5mm] (replay.east)--(13.85,.30)--(13.85,1.20);
\end{tikzpicture}}
\caption{Two complementary training-data streams. Natural text supplies CPT and material for derived supervision. Public resources join constructed examples through curation; replay enters the final SFT mixture separately. Dashed arrows identify derived supervision and replay inputs. Construction includes context-guided generation, human verification, normalization, and literary curation, as described in the text.}
\label{fig:datarecipes}
\end{figure}
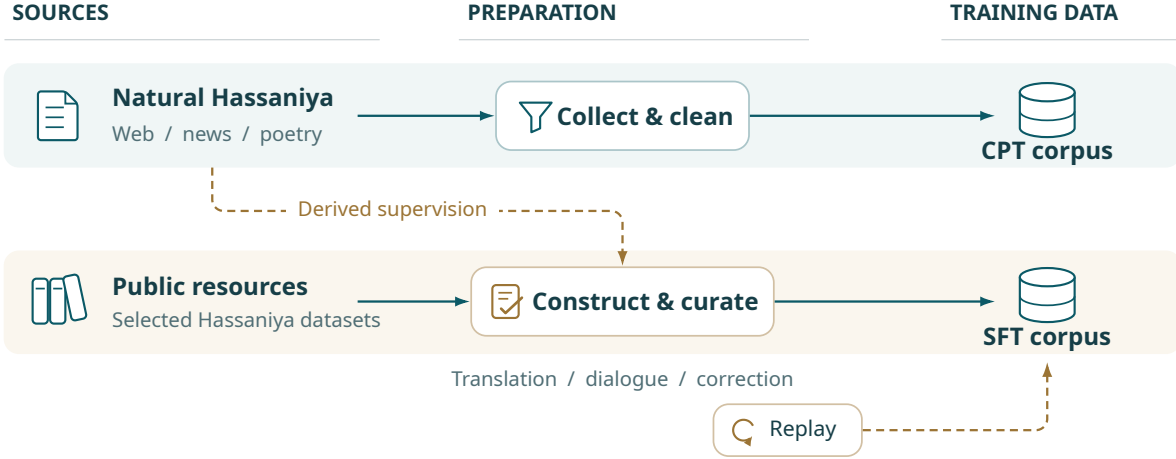

Beyond integrating public resources, Enemray relies on several explicit construction recipes designed around the forms in which Hassaniya is naturally available. These recipes preserve the distinction between naturally occurring text, machine-assisted derived supervision, and deliberately constructed normalization data.
Figure~\ref{fig:datarecipes} summarizes the main recipe families and how they feed either the continual-pretraining stream or the Hassaniya supervised mixture.

\paragraph{Natural web text for continual pretraining.}
The continual-pretraining stream is built primarily through web collection from many Mauritanian online sources. We collect naturally occurring Hassaniya posts, social narratives, stories, articles, news, and related text, then organize the retained material as document-level causal-language-modeling examples. This stream is intentionally kept separate from the supervised public datasets in Table~\ref{tab:ecosystem}: its role is exposure to the language distribution itself rather than instruction following. Social posts and narratives form the largest component because they provide substantial naturally occurring colloquial Hassaniya, while journalism, long-form articles, cultural writing, and literary material broaden register coverage.

\paragraph{Context-guided pseudo-parallel translation.}
Web posts and stories collected for continual pretraining are also used to build supervised translation data. Each document is split into short and medium Hassaniya segments, which become the source side of the translation pairs. A short segment taken on its own is often ambiguous: it may refer to people or events mentioned earlier, omit information that is clear from the surrounding text, or use a local expression whose meaning depends on the situation. For this reason, Gemini \cite{gemini} does not translate segments in isolation. It receives the full source document together with the selected segment, and produces an English rendering of that segment only, using the document to resolve the narrative situation, referents, and dialectal expressions. Because the English side is produced by a model, the resulting pairs are treated as \emph{machine-generated pseudo-translations} rather than human-authored gold translations. Before inclusion, each rendering is checked by a human against the Hassaniya segment and its context.

\paragraph{Context-guided conversation construction.}
User--assistant conversations are built from the same natural Hassaniya documents. Generating dialogue from an isolated sentence tends to produce generic exchanges that are only loosely related to the source. Instead, Gemini receives the source document and writes conversational turns grounded in its content, so that the exchange reflects the situations, people, and topics that appear in the original text. As with the translation data, each candidate conversation is verified by a human before it enters the supervised mixture. A single source document can therefore contribute both pseudo-parallel translation pairs and contextual assistant conversations, while both types of supervision remain tied to naturally occurring Hassaniya.

\paragraph{Correction and Hassaniya normalization.}
Correction data is constructed in the reverse direction: a clean Hassaniya sentence is kept as the assistant target, while a controlled noisy version is presented as the user input. The corruption pipeline includes rule-based character, spacing, punctuation, spelling, and local word-order noise; machine-translation (MT) and back-translation noise that can introduce translationese or less-natural phrasing; and conservative substitution of selected Hassaniya forms with Modern Standard Arabic (MSA) or Moroccan Darija alternatives. The goal is broader than spelling repair: the model is trained to normalize noisy, mixed, or model-like text back into natural Hassaniya while preserving meaning. Generated correction examples are built only from the training split of the clean translation targets; held-out translation examples and Alexandria translation evaluation data are excluded from this augmentation.

\paragraph{Poetry and literary curation.}
Because poetry and oral expression are central to Hassaniya cultural and linguistic coverage, we also select substantial Hassaniya poetic and literary material from additional sources rather than relying only on generic Arabic corpora. Raw literary text contributes language exposure where appropriate, while structured examples are used to teach supervised literary and cultural behavior, including Hassaniya poetic forms such as Aagel, Lghna, and Tebraa. This provides lexical, stylistic, figurative, and culturally grounded patterns that are comparatively sparse in ordinary web prose.

\subsection{Continual-pretraining corpus}
\label{sec:cptdata}

The continual-pretraining corpus exposes the base model to natural text before assistant specialization. The clean corpus contains 11,629 documents totaling approximately 20.9 million Gemma 4 tokens (9.5 million words). Documents are typically long-form, with a median length of 581 tokens. Its source domains are summarized in Table~\ref{tab:cptdata_table}.

\begin{table}[H]
\centering
\caption{Source domains of the continual-pretraining corpus.}
\label{tab:cptdata_table}
\small
\begin{tabularx}{\textwidth}{@{}>{\raggedright\arraybackslash}p{0.42\textwidth} X@{}}
\toprule
Source domain & Contribution \\
\midrule
Standalone social-media narratives & Colloquial writing and everyday discourse \\
Reconstructed long-form narratives & Extended narrative and contextual continuity \\
Mauritanian news media & Contemporary events and expository language \\
Mauritanian long-form articles & Longer factual and topical discussion \\
Cultural essays & Local knowledge and cultural expression \\
Poetry and language-learning texts & Literary forms and linguistic explanation \\
\bottomrule
\end{tabularx}
\end{table}

The mixture is intentionally register-diverse. Social narratives provide everyday dialectal discourse, dialogue, interpersonal relations, narrative progression, colloquial lexical choice, and naturally varying orthography. Journalism and long-form articles extend coverage toward factual, expository, and more formal registers. Cultural essays and language-learning material add local knowledge and pedagogical structure, while poetry introduces expressive forms that are culturally salient in Hassaniya and difficult to recover from general Arabic corpora alone \cite{unesco_t_heydinn}.

\subsection{Hassaniya supervised corpus}
\label{sec:sftdata}

The Hassaniya supervised data comprise 46,259 conversations and 49,964 assistant turns, with approximately 3.0 million tokens of user and assistant content, of which 1.5 million tokens form the assistant responses used as training targets. Selected public datasets contribute high-value forms of supervision that are difficult to recover reliably from raw text alone. DAH contributes English--Hassaniya parallel examples; the Mauritanian portion of Alexandria adds multi-domain conversational translation; PALM contributes country-specific cultural and dialectal instructions; selected AI-for-RIM customer-support dialogues add practical multi-turn interaction; Amthal Hassaniya contributes proverb explanation and idiomatic grounding; and Hassaniya scripture parallels add further cross-lingual lexical and syntactic coverage \cite{hassania_dah,elmekki2026alexandria,alwajih2025palm,aiforrim,amthalhassaniya,scriptureearthmey,bibleparallel}.

Beyond these public sources, Enemray data extend supervision across five capability families: \textbf{cultural and contextual grounding}, including Mauritanian heritage, history, local knowledge, and question answering; \textbf{poetry and oral tradition}, including Aagel, Lghna, and Tebraa; \textbf{situational conversation}, including role-play and everyday social scenarios; \textbf{cross-lingual communication}, including English--Hassaniya dialogue and translation; and \textbf{linguistic refinement}, including grammar, spelling, phrasing, lexical choice, correction, and stylistic naturalness.

\subsection{Curation, separation, and supervision}
\label{sec:datacuration}

The data pipeline is organized around the objective of each training stage. Natural text is used for continual pretraining, where it is presented as documents under a causal language-modeling objective, whereas instruction, translation, proverb, dialogue, and parallel-text resources are reserved for supervised post-training. This separation distinguishes learning the distribution of the language from learning how to respond within an interaction.

In the supervised stage, conversations are represented in the native conversational format of the model, and the training signal is restricted to assistant responses. Each assistant turn forms its own training example, with the preceding dialogue serving as context, so that the model learns to produce responses rather than to reproduce prompts. The general-capability replay set is kept separate from the Hassaniya data and is added only to the final supervised mixture, where it serves as a retention signal (Section~\ref{sec:training}).

Cleaning is deliberately conservative. For the continual-pretraining corpus, it removes invisible formatting characters while preserving every document, including dialectal spelling and informal punctuation. For the supervised corpus, it removes exact duplicate conversations and residual markup from generated responses. These steps ensure structural consistency across the data.

Because supervision is built from the same body of Hassaniya literature, some poems appear both as documents in the continual-pretraining corpus and as responses in the supervised poetry data. The model therefore encounters these texts first as natural language and subsequently as targets of literary generation.

\section{Training Methodology}
\label{sec:training}

Enemray is trained in three stages that separate language acquisition, language transfer, and supervised post-training (Figure~\ref{fig:pipeline}). The configuration of each training stage is summarized in Table~\ref{tab:trainingcfg}.

\subsection{Stage I: Continual pretraining}

The base checkpoint is adapted to Hassaniya under the causal language-modeling objective, with low-rank adapters restricted to the layers and projections described in Section~\ref{sec:model}. Documents from the continual-pretraining corpus are separated by an end-of-sequence token and packed into sequences of 2,048 tokens. Training runs for two epochs, during which the model processes approximately 42 million tokens. Optimization uses 8-bit AdamW with a learning rate of $5\times10^{-5}$, linear decay after a short warmup, and an effective batch size of 16 sequences.

\subsection{Stage II: Language transfer}

The trained language adapter is transferred to the instruction-tuned checkpoint by scaling its update with $\gamma=0.7$ and merging it into the model weights, as defined in Equation~\ref{eq:transfer}. The merge is performed in full precision. On a reference input, the merged model reproduces the logits of the unmerged adapted model to within a maximum absolute difference of $1.7\times10^{-4}$, confirming that merging preserves the behavior of the adapted model.

\subsection{Stage III: Supervised post-training}

The grafted checkpoint is then trained on the supervised mixture described in Section~\ref{sec:sftdata}, with adaptation extended to all attention and feed-forward projections across the full depth of the decoder (Section~\ref{sec:model}). Each assistant turn forms a separate training example, in which the assistant response is the target and the preceding dialogue serves as context. Let $T(x)$ denote the target token positions of an example $x$. The training objective is
\begin{equation}
\mathcal{L}_{\mathrm{SFT}}(\theta)
=-\frac{1}{\sum_x |T(x)|}
\sum_x\sum_{t\in T(x)}\log p_\theta(x_t\mid x_{<t}),
\end{equation}
so that prompt and padding tokens do not contribute to the loss.

Conversations are rendered in the native Gemma 4 conversational format with the reasoning mode disabled, and sequences are limited to 8,192 tokens. When a dialogue exceeds this length, the earliest context is removed so that the target response is always retained in full. Training runs for two epochs with AdamW at a learning rate of $5\times10^{-5}$ and an effective batch size of 16, with examples sampled in random order.

The supervised mixture also contains the general-capability replay set, which covers reasoning, coding, multilingual assistance, and instruction following. Replay accounts for approximately 17\% of the supervised target tokens and maintains exposure to general assistant behavior while the model specializes in Hassaniya, following evidence that replay of model-generated responses reduces forgetting during continual fine-tuning \cite{chen2026opr}.

\subsection{Training setup}

All stages are trained on a single NVIDIA H100 80\,GB GPU. Model computation uses BF16 precision, while the adapter merge in Stage~II is performed in FP32 before the grafted model is stored in BF16. The complete multimodal checkpoint is loaded throughout, and the native Gemma 4 processor and conversational format are retained \cite{gemma4transformers}.

\begin{table}[t]
\centering
\caption{Training configuration of the continual-pretraining and supervised post-training stages.}
\label{tab:trainingcfg}
\small
\begin{tabularx}{\textwidth}{@{}>{\bfseries}l X X@{}}
\toprule
Setting & Continual pretraining & Supervised post-training \\
\midrule
Maximum sequence length & 2,048 & 8,192 \\
Effective batch size & 16 & 16 \\
Gradient accumulation & 1 & 1 \\
Epochs & 2 & 2 \\
Learning rate & $5\times10^{-5}$ & $5\times10^{-5}$ \\
Learning-rate schedule & Linear & Linear \\
Warmup fraction & 0.03 & 0.03 \\
Weight decay & 0.01 & 0.01 \\
Gradient clipping & 1.0 & 1.0 \\
Optimizer & 8-bit AdamW & AdamW \\
LoRA rank / scaling parameter & 16 / 32 & 16 / 32 \\
LoRA dropout & 0 & 0 \\
Adapted weight matrices & 72 in 12 layers & 294 in 42 layers \\
Loss & Causal language modeling & Assistant responses \\
Random seed & 42 & 42 \\
\bottomrule
\end{tabularx}
\end{table}

\section{Evaluation}
\label{sec:evaluation}

We evaluate Enemray along three axes: translation between English and Hassaniya, detection and classification of errors in Hassaniya machine translation, and retention of general capabilities after adaptation. All metrics are computed automatically against fixed references or executable tests, without human or model-based judges. Table~\ref{tab:evalsetup} summarizes the experimental setup.

\begin{table}[H]
\centering
\caption{Evaluation setup. All samples are drawn once with a fixed random seed and shared across models.}
\label{tab:evalsetup}
\small
\begin{tabularx}{\textwidth}{@{}>{\raggedright\arraybackslash}p{0.2\textwidth} >{\raggedright\arraybackslash}X r >{\raggedright\arraybackslash}p{0.2\textwidth}@{}}
\toprule
Experiment & Data & Samples & Metric \\
\midrule
Translation & Alexandria (Mauritania), English $\rightarrow$ Hassaniya and Hassaniya $\rightarrow$ English & 2,000 & spBLEU, chrF++ \\
Error detection & AlexandriaX Subtask 3 (Mauritania), LQM-annotated error spans & 274 & Exact match F1, overlap F1, class F1 \\
Math reasoning & GSM8K \cite{cobbe2021gsm8k} & 500 & Exact match \\
Knowledge & MMLU-Pro \cite{wang2024mmlupro} & 500 & Accuracy \\
Code generation & HumanEval \cite{chen2021humaneval} & 164 & pass@1 \\
Function calling & BFCL \cite{patil2025bfcl} & 500 & Call accuracy \\
\bottomrule
\end{tabularx}
\end{table}

\paragraph{Models and decoding.}
Enemray is compared with its underlying instruction-tuned model, Gemma 4 E4B-it, so that differences reflect the adaptation pipeline. For translation and error detection, we additionally include three proprietary and open models accessed through OpenRouter: GPT-5.6 Luna, Gemini 3.5 Flash Lite, and Qwen 3.8 Flash. For translation, we also report Amalaz, a pair of small direction-specific MarianMT models for English $\rightarrow$ Hassaniya and Hassaniya $\rightarrow$ English, which translate the source sentence directly with their default beam-search decoding. Gemma 4 E4B-it and Enemray are decoded greedily with the reasoning mode disabled. The external models are queried with a fixed seed, a low reasoning effort, and the provider's default sampling configuration. Within each experiment, every model receives the same prompt.

\subsection{Translation}

Translation is evaluated on the Mauritanian portion of Alexandria \cite{elmekki2026alexandria}, a human-translated and reviewed collection of multi-domain dialogues. From its test split, we sample 1,000 parallel dialogue turns covering all 11 domains and evaluate each turn in both directions, yielding 1,000 English $\rightarrow$ Hassaniya and 1,000 Hassaniya $\rightarrow$ English examples. Each model receives a neutral translation instruction specifying the source and target languages and asking for the translation only. Outputs are scored against the reference translation with spBLEU, computed with the FLORES-200 SentencePiece tokenizer \cite{nllb2022}, and chrF++ \cite{popovic2017chrf}, both implemented in sacreBLEU \cite{post2018sacrebleu}.

\begin{table}[t]
\centering
\caption{Translation results on the Alexandria (Mauritania) test set, 1,000 examples per direction. Amalaz denotes the direction-specific MarianMT models amalaz-en2ha-micro and amalaz-ha2en-micro. The best score in each row is shown in bold.}
\label{tab:translation}
\small
\setlength{\tabcolsep}{4pt}
\begin{tabular}{@{}l l c c c c c c@{}}
\toprule
 & \textbf{Metric} & \textbf{Amalaz} & \textbf{Gemma 4} & \textbf{GPT-5.6} & \textbf{Gemini 3.5} & \textbf{Qwen 3.8} & \textbf{Enemray} \\
 & & \textbf{micro} & \textbf{E4B-it} & \textbf{Luna} & \textbf{Flash Lite} & \textbf{Flash} & \\
\midrule
English $\rightarrow$ Hassaniya & spBLEU & 5.74 & 8.83 & 8.61 & 5.32 & 8.30 & \textbf{12.64} \\
 & chrF++ & 21.96 & 26.09 & 26.14 & 21.59 & 25.34 & \textbf{29.42} \\
\midrule
Hassaniya $\rightarrow$ English & spBLEU & 13.22 & 21.94 & 26.95 & 28.31 & 20.62 & \textbf{29.52} \\
 & chrF++ & 32.76 & 42.91 & 48.76 & \textbf{49.93} & 44.18 & 48.99 \\
\bottomrule
\end{tabular}
\end{table}

\paragraph{Results.}
Table~\ref{tab:translation} and Figure~\ref{fig:translation} report the translation results. Enemray achieves the highest English $\rightarrow$ Hassaniya scores on both metrics, improving over Gemma 4 E4B-it by 3.8 spBLEU and 3.3 chrF++. In the Hassaniya $\rightarrow$ English direction, Enemray obtains the highest spBLEU, while Gemini 3.5 Flash Lite achieves the highest chrF++.

\subsection{Error detection}

The error-detection experiment measures whether a model can recognize what is wrong in a Hassaniya translation, rather than produce or repair one. It follows Subtask 3 of the AlexandriaX-2026 shared task \cite{alexandriax2026}, in which a system receives an English sentence and its machine translation into an Arabic dialect, identifies the erroneous spans in the translation, and assigns each span an error category. The underlying data were built with the LQM framework \cite{magdy2026lqm}: English sentences derived from conversational dialect transcripts were translated by six Arabic-aware large language models, and native-speaker expert annotators marked the erroneous spans and labeled them with one of six linguistically grounded categories: sociolinguistics, pragmatics, semantics, morphosyntax, orthography and writing conventions, and graphetics. We use all 274 English $\rightarrow$ Mauritanian Hassaniya examples with released annotations, comprising the 244 training and 30 development examples of the shared task, which together contain 390 annotated error spans.

Each model receives the English source, the translation, and the list of the six categories with their LQM definitions, and returns the erroneous spans with their categories in JSON format. Character offsets are obtained by locating each returned span in the translation. Predictions are scored with the official shared-task scorer, which reports exact match F1, where a predicted span must match the gold character offsets exactly; overlap F1, computed over the characters marked as erroneous; and class F1, where predicted and gold spans are paired by their overlap and a pair counts as correct when the categories agree. Following the shared task, the overall score is the average of overlap F1 and class F1. We report results on all 274 examples.

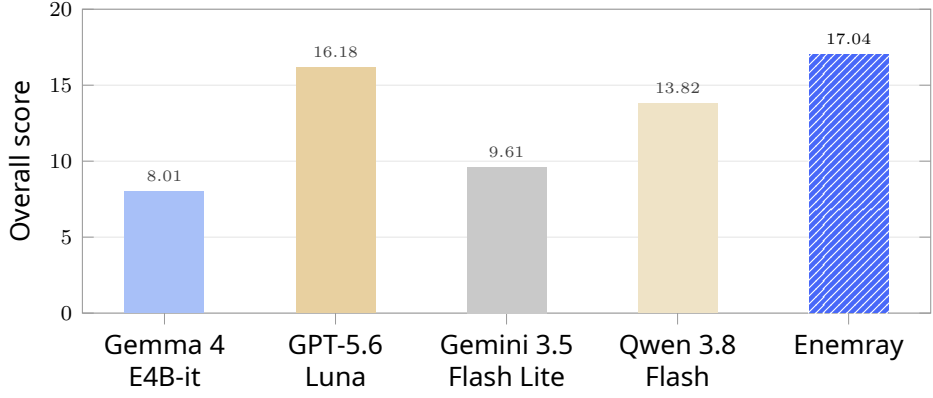
\begin{figure}[t]
\centering
% Error-detection overall score bar chart (same style as translation_results.tex).
% Scores: eval/results/error_detection (official AlexandriaX overall score x 100).
\definecolor{BarOurs}{HTML}{4868F8}
\definecolor{BarBase}{HTML}{A8C0F8}
\definecolor{BarGPT}{HTML}{E8D0A0}
\definecolor{BarGemini}{HTML}{C9C9C9}
\definecolor{BarQwen}{HTML}{EFE3C6}
\begin{tikzpicture}
\begin{axis}[
  width=0.8\textwidth,
  height=5.6cm,
  ybar,
  bar width=30pt,
  bar shift=0pt,
  ymin=0, ymax=20,
  ylabel={Overall score},
  ylabel style={font=\sffamily\small},
  symbolic x coords={Gemma,GPT,Gemini,Qwen,Enemray},
  xtick={Gemma,GPT,Gemini,Qwen,Enemray},
  xticklabels={{Gemma 4\\E4B-it},{GPT-5.6\\Luna},{Gemini 3.5\\Flash Lite},{Qwen 3.8\\Flash},{Enemray}},
  xticklabel style={font=\sffamily\small, align=center},
  enlarge x limits=0.12,
  ymajorgrids,
  grid style={draw=black!10},
  axis line style={draw=black!40},
  tick style={draw=black!40},
  xtick pos=lower,
  yticklabel style={font=\sffamily\scriptsize},
  nodes near coords,
  nodes near coords style={font=\sffamily\fontsize{6.5}{7}\selectfont, text=black!75,
    /pgf/number format/fixed, /pgf/number format/fixed zerofill, /pgf/number format/precision=2},
  clip=false,
]
\addplot[fill=BarBase, draw=none] coordinates {(Gemma,8.01)};
\addplot[fill=BarGPT, draw=none] coordinates {(GPT,16.18)};
\addplot[fill=BarGemini, draw=none] coordinates {(Gemini,9.61)};
\addplot[fill=BarQwen, draw=none] coordinates {(Qwen,13.82)};
\addplot[fill=BarOurs, draw=none, postaction={pattern=north east lines, pattern color=white},
  nodes near coords style={font=\sffamily\fontsize{6.5}{7}\selectfont\bfseries, text=black}]
  coordinates {(Enemray,17.04)};
\end{axis}
\end{tikzpicture}
\caption{Overall error-detection score on the Mauritanian portion of AlexandriaX Subtask 3, the average of overlap F1 and class F1 (multiplied by 100).}
\label{fig:errordetection}
\end{figure}

\begin{table}[t]
\centering
\caption{Error-detection results on the Mauritanian portion of AlexandriaX Subtask 3 (274 examples, 390 gold error spans). Scores are multiplied by 100, and the best score in each metric row is shown in bold. The lower rows describe prediction behavior; in the gold annotation, erroneous spans cover 19.5\% of each translation on average.}
\label{tab:errordetection}
\small
\setlength{\tabcolsep}{4.5pt}
\begin{tabular}{@{}l c c c c c@{}}
\toprule
 & \textbf{Gemma 4} & \textbf{GPT-5.6} & \textbf{Gemini 3.5} & \textbf{Qwen 3.8} & \textbf{Enemray} \\
 & \textbf{E4B-it} & \textbf{Luna} & \textbf{Flash Lite} & \textbf{Flash} & \\
\midrule
Overall & 8.01 & 16.18 & 9.61 & 13.82 & \textbf{17.04} \\
Exact match F1 & 3.39 & 9.97 & 6.37 & \textbf{10.03} & 2.79 \\
Overlap F1 & 12.21 & 22.13 & 13.55 & 20.39 & \textbf{27.58} \\
Class F1 & 3.81 & \textbf{10.23} & 5.66 & 7.24 & 6.51 \\
\midrule
Examples reported as error-free & 227 & 60 & 174 & 79 & 19 \\
Translation marked as erroneous (\%) & 6.9 & 23.1 & 8.2 & 16.2 & 46.9 \\
\bottomrule
\end{tabular}
\end{table}

\paragraph{Results.}
Table~\ref{tab:errordetection} and Figure~\ref{fig:errordetection} report the error-detection results. The task remains difficult for all models. Adaptation markedly changes the behavior of the base model: Gemma 4 E4B-it reports no error for 227 of the 274 translations, all of which contain at least one annotated error, whereas Enemray reports errors for all but 19 and obtains the highest overall and overlap scores. Its spans are, however, considerably broader than the gold annotations, covering 46.9\% of each translation on average compared with 19.5\%, which yields the lowest exact match F1. Category assignment is the main weakness across models: although sociolinguistic errors account for 81\% of the gold spans, the models predominantly label errors as semantic or morphosyntactic.

\subsection{Capability retention}

Capability retention compares Enemray with Gemma 4 E4B-it on four established benchmarks, using the same prompts, no system prompt, and greedy decoding for both models.

\paragraph{Mathematical reasoning.} We sample 500 problems from the GSM8K test set \cite{cobbe2021gsm8k}. The model reasons step by step and states its final answer in a fixed format, and a response is correct when the extracted number equals the reference answer.

\paragraph{Knowledge and reasoning.} We sample 500 questions from the MMLU-Pro test set \cite{wang2024mmlupro}, stratified by subject category. Each question has up to ten answer options; the model reasons step by step and concludes with the selected letter, which is extracted following the official MMLU-Pro evaluation. We report accuracy.

\paragraph{Code generation.} All 164 HumanEval problems \cite{chen2021humaneval} are used. The model completes each function from its signature and docstring, and the completed program is executed against the official unit tests. We report pass@1.

\paragraph{Function calling.} We sample 500 single-turn problems from the Berkeley Function Calling Leaderboard \cite{patil2025bfcl}: 200 simple, 100 multiple-function, 100 parallel, and 100 irrelevance cases. Tool definitions are provided through the native Gemma 4 function-calling format, and the generated calls are verified with abstract-syntax-tree matching following the official BFCL criteria, which check the function name, required parameters, and parameter values. In the irrelevance category, a response is correct when the model makes no function call.

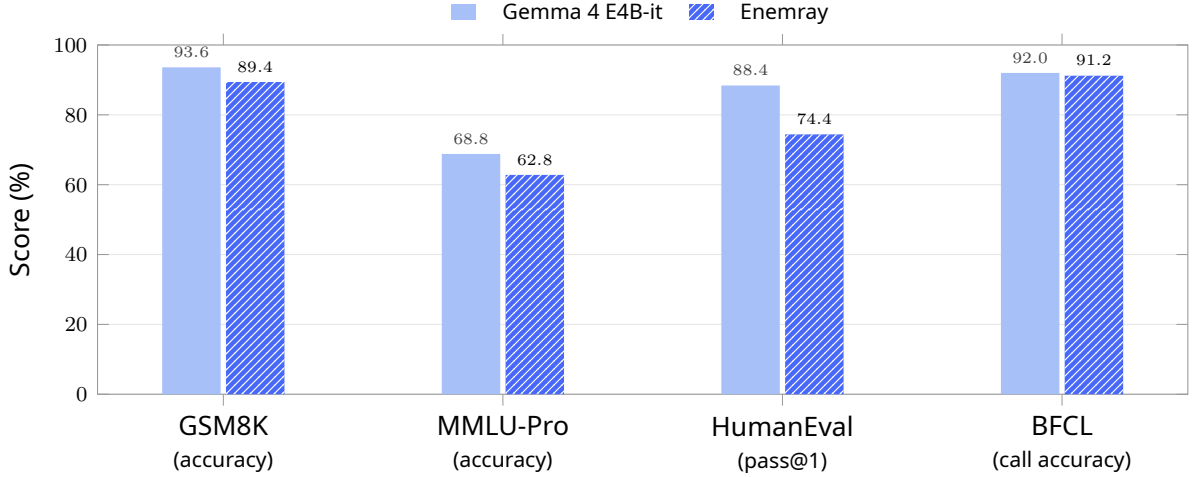
\begin{figure}[t]
\centering
% Capability retention grouped bar chart (same style as translation_results.tex).
% Scores: eval/results/benchmarks.
\definecolor{BarOurs}{HTML}{4868F8}
\definecolor{BarBase}{HTML}{A8C0F8}
\begin{tikzpicture}
\begin{axis}[
  width=\textwidth,
  height=6.2cm,
  ybar=2pt,
  bar width=22pt,
  ymin=0, ymax=100,
  ylabel={Score (\%)},
  ylabel style={font=\sffamily\small},
  symbolic x coords={GSM8K,MMLU-Pro,HumanEval,BFCL},
  xtick=data,
  xticklabels={{GSM8K\\\scriptsize(accuracy)},{MMLU-Pro\\\scriptsize(accuracy)},{HumanEval\\\scriptsize(pass@1)},{BFCL\\\scriptsize(call accuracy)}},
  xticklabel style={font=\sffamily\small, align=center},
  enlarge x limits=0.15,
  ymajorgrids,
  grid style={draw=black!10},
  axis line style={draw=black!40},
  tick style={draw=black!40},
  yticklabel style={font=\sffamily\scriptsize},
  nodes near coords,
  nodes near coords style={font=\sffamily\fontsize{6.5}{7}\selectfont, text=black!75,
    /pgf/number format/fixed, /pgf/number format/fixed zerofill, /pgf/number format/precision=1},
  legend style={draw=none, font=\sffamily\scriptsize, legend columns=2, column sep=8pt,
    at={(0.5,1.03)}, anchor=south},
  legend image code/.code={\draw[#1, draw=none] (0cm,-0.08cm) rectangle (0.32cm,0.12cm);},
  clip=false,
]
\addplot[fill=BarBase, draw=none] coordinates {(GSM8K,93.6) (MMLU-Pro,68.8) (HumanEval,88.4) (BFCL,92.0)};
\addlegendentry{Gemma 4 E4B-it}
\addplot[fill=BarOurs, draw=none, postaction={pattern=north east lines, pattern color=white},
  nodes near coords style={font=\sffamily\fontsize{6.5}{7}\selectfont\bfseries, text=black}]
  coordinates {(GSM8K,89.4) (MMLU-Pro,62.8) (HumanEval,74.4) (BFCL,91.2)};
\addlegendentry{Enemray}
\end{axis}
\end{tikzpicture}
\caption{Capability retention of Enemray compared with Gemma 4 E4B-it on GSM8K, MMLU-Pro, HumanEval, and BFCL.}
\label{fig:retention}
\end{figure}

\begin{figure}[t]
\centering
% HumanEval pass@1 with and without Enemray's reasoning mode (same style as retention_results.tex).
% Scores: eval/results/benchmarks/humaneval and humaneval_thinking.
\definecolor{BarOurs}{HTML}{4868F8}
\definecolor{BarBase}{HTML}{A8C0F8}
\definecolor{BarOursThink}{HTML}{2A3FB0}
\begin{tikzpicture}
\begin{axis}[
  width=0.62\textwidth,
  height=5.2cm,
  ybar,
  bar width=34pt,
  bar shift=0pt,
  ymin=0, ymax=100,
  ylabel={HumanEval pass@1 (\%)},
  ylabel style={font=\sffamily\small},
  symbolic x coords={Gemma,Enemray,EnemrayThink},
  xtick={Gemma,Enemray,EnemrayThink},
  xticklabels={{Gemma 4 E4B-it},{Enemray},{Enemray\\(reasoning)}},
  xticklabel style={font=\sffamily\small, align=center},
  enlarge x limits=0.25,
  ymajorgrids,
  grid style={draw=black!10},
  axis line style={draw=black!40},
  tick style={draw=black!40},
  xtick pos=lower,
  yticklabel style={font=\sffamily\scriptsize},
  nodes near coords,
  nodes near coords style={font=\sffamily\fontsize{6.5}{7}\selectfont, text=black!75,
    /pgf/number format/fixed, /pgf/number format/fixed zerofill, /pgf/number format/precision=1},
  clip=false,
]
\addplot[fill=BarBase, draw=none] coordinates {(Gemma,88.4)};
\addplot[fill=BarOurs, draw=none, postaction={pattern=north east lines, pattern color=white},
  nodes near coords style={font=\sffamily\fontsize{6.5}{7}\selectfont\bfseries, text=black}]
  coordinates {(Enemray,74.4)};
\addplot[fill=BarOursThink, draw=none, postaction={pattern=north east lines, pattern color=white},
  nodes near coords style={font=\sffamily\fontsize{6.5}{7}\selectfont\bfseries, text=black}]
  coordinates {(EnemrayThink,87.8)};
\end{axis}
\end{tikzpicture}
\caption{HumanEval pass@1 of Enemray with and without its reasoning mode, compared with Gemma 4 E4B-it. Without reasoning, both models are decoded greedily; with reasoning, Enemray is sampled at temperature 0.8 with up to 8,192 new tokens.}
\label{fig:humanevalthinking}
\end{figure}
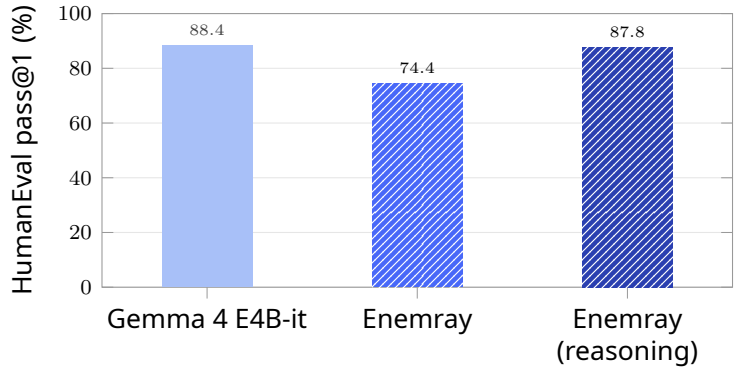

\paragraph{Results.}
Figure~\ref{fig:retention} reports the capability retention results. Enemray retains most of the general capabilities of Gemma 4 E4B-it, with some reduction after adaptation. Function calling is largely unchanged, with BFCL accuracy of 91.2\% compared with 92.0\%. On GSM8K and MMLU-Pro, Enemray reaches 89.4\% and 62.8\%, retaining 95.5\% and 91.3\% of the base model's accuracy. The largest reduction appears in code generation, where HumanEval pass@1 decreases from 88.4\% to 74.4\%. Enemray's solutions tend to be shorter and to implement the function directly, whereas the base model more often works through the problem in code comments before implementing it. This suggests that much of the gap reflects how the model approaches the problem rather than a loss of coding knowledge. Consistent with this, when Enemray is evaluated with its reasoning mode enabled (sampling temperature 0.8, up to 8,192 new tokens), its HumanEval pass@1 rises to 87.8\%, nearly matching the 88.4\% of the base model without reasoning (Figure~\ref{fig:humanevalthinking}). Although replay accounts for approximately 17\% of the supervised target tokens, it consists of only 273 conversations, and coding tasks form a small subset of them. Increasing the number and diversity of policy-generated replay conversations, with broader coverage of coding and reasoning tasks, is a direct way to further improve retention in future versions of Enemray.

\section{Limitations and Release}
\label{sec:limitations}

\subsection{Limitations}

Enemray is trained on a collection that emphasizes natural narratives, social writing, Mauritanian journalism, cultural material, and targeted assistant tasks. Although this collection spans a broad range of registers, coverage of specialized domains such as scientific, legal, and professional writing remains limited, and competence in these domains is correspondingly weaker. Hassaniya is also primarily an oral variety with considerable variation in spelling, code-switching, and proximity to Standard Arabic \cite{tainecheikh2020}. Model outputs may therefore alternate between Hassaniya and Standard Arabic forms, and judgments of dialectal naturalness, cultural appropriateness, and poetic form ultimately require native-speaker expertise.

The present adaptation is restricted to text. Although the underlying Gemma 4 E4B checkpoint supports image and audio input, the vision and audio components are not adapted, and the multimodal behavior of Enemray is outside the scope of this report.

\subsection{Release of the current checkpoint}

The Enemray checkpoint described in this report will not be publicly released. The base Gemma 4 E4B model is distributed under the Apache~2.0 license \cite{gemma4base} and does not itself restrict the release of derived models. The constraint arises instead from the supervised training data. Several third-party resources incorporated into the supervised mixture are distributed under terms that limit their use and redistribution. Alexandria is released under a Creative Commons Attribution--NonCommercial license, with access conditioned on an agreement to non-commercial use \cite{elmekki2026alexandria}, and PALM is likewise distributed under non-commercial terms \cite{alwajih2025palm}. Other resources in the mixture do not state explicit licensing terms, and their use in a publicly distributed model would require permission from the respective rights holders. Because model weights trained on these resources cannot be released without such permissions, the current checkpoint is not made publicly available.

\subsection{Toward a releasable Enemray}

We are actively working toward a version of Enemray that can be released openly. This effort follows two directions. First, resources whose licensing terms require permission for redistribution or derivative use are being removed from the training mixture. Second, we are developing and curating locally sourced Hassaniya data to replace these resources and to further extend coverage of Hassaniya language, culture, and interaction. The resulting corpus will be used to train a more capable Enemray model, which we intend to release publicly.

Beyond changes to the data, the next version will extend the training pipeline beyond supervised fine-tuning. We plan to incorporate reinforcement learning from human feedback, with preferences provided by native Hassaniya speakers, to better align the model's responses with the expectations of Hassaniya users in terms of naturalness, cultural appropriateness, and helpfulness. We also plan to simulate native Hassaniya-speaking users, so that the model can be trained and assessed on realistic multi-turn interactions that reflect how Hassaniya speakers actually phrase requests and conduct conversations. Finally, the next version is intended to extend Enemray beyond text by adapting both its audio and image capabilities, allowing the model to understand spoken Hassaniya and to engage with visual content in Hassaniya.

\section{Additional Examples}
\label{sec:examples}

Table~\ref{tab:examples} presents additional examples of Enemray outputs covering everyday advice, English $\rightarrow$ Hassaniya translation, and poetry generation.

\begin{table}[H]
\centering
\caption{Additional examples of Enemray outputs. Outputs are unedited and were generated with greedy decoding and the reasoning mode disabled.}
\label{tab:examples}
\includegraphics[width=\textwidth]{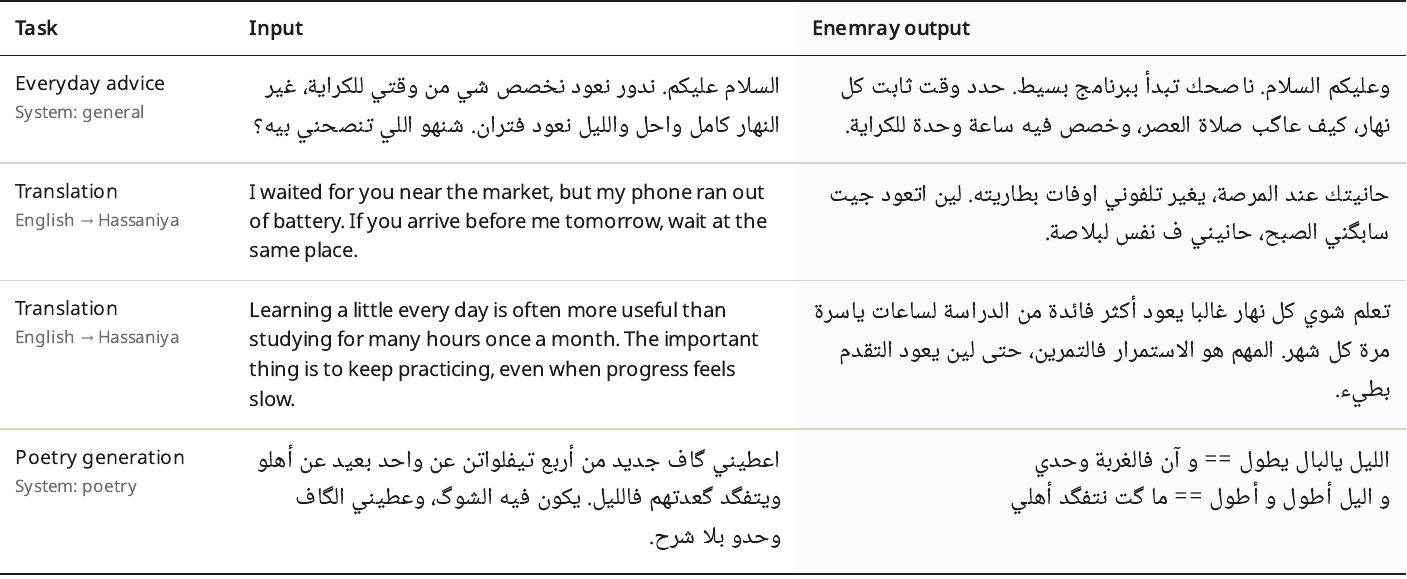}
\end{table}

\section{Conclusion}
\label{sec:conclusion}

This report presented Enemray, a Hassaniya-centric language model built on Gemma 4 E4B. Enemray approaches the adaptation of a capable instruction-tuned model to an underrepresented language as a stability--plasticity problem and addresses it by separating language acquisition from behavioral specialization. Layer-selective continual pretraining learns a compact Hassaniya update from natural text, this update is transferred into the instruction-tuned model through parameter-space composition, and supervised post-training with general-capability replay shapes Hassaniya interaction while maintaining broader assistant behavior. Supporting this pipeline, we assembled a continual-pretraining corpus of natural and long-form Hassaniya text and a supervised corpus that combines selected public resources with a larger body of Enemray-constructed data covering cultural knowledge, poetry, conversation, translation, and linguistic refinement.

The evaluation indicates that this design yields clear gains in Hassaniya while largely preserving general capabilities. Enemray achieves the strongest English $\rightarrow$ Hassaniya translation among the evaluated models, including the proprietary systems, and the highest spBLEU in the Hassaniya $\rightarrow$ English direction. It also obtains the highest overall score on Mauritanian error detection, although precise span localization and error categorization remain difficult for all models. On general benchmarks, Enemray retains most of the capabilities of its instruction-tuned base model: function calling is largely unchanged, mathematical reasoning and knowledge show modest reductions, and the reduction in code generation is almost entirely recovered when the reasoning mode is enabled.

Several directions remain open. The next version of Enemray will replace resources with restrictive licensing terms by locally sourced Hassaniya data so that the model can be released openly. It will also increase the amount and coverage of policy-generated replay, incorporate reinforcement learning from human feedback with native Hassaniya speakers and simulated Hassaniya-speaking users, and extend adaptation to audio and images. We hope that Enemray and the methodology documented in this report contribute to making Hassaniya a first-class language for modern language technology.

\bibliographystyle{unsrtnat}
\bibliography{references}
\end{document}